\documentclass[letterpaper]{article} % DO NOT CHANGE THIS
\usepackage{aaai25}  % DO NOT CHANGE THIS
\usepackage{times}  % DO NOT CHANGE THIS
\usepackage{helvet}  % DO NOT CHANGE THIS
\usepackage{courier}  % DO NOT CHANGE THIS
\usepackage[hyphens]{url}  % DO NOT CHANGE THIS
\usepackage{graphicx} % DO NOT CHANGE THIS
\usepackage{natbib}  % DO NOT CHANGE THIS AND DO NOT ADD ANY OPTIONS TO IT
\usepackage{caption} % DO NOT CHANGE THIS AND DO NOT ADD ANY OPTIONS TO IT
\usepackage{algorithm}
\usepackage{algorithmic}

\usepackage{newfloat}
\usepackage{listings}
\DeclareCaptionStyle{ruled}{labelfont=normalfont,labelsep=colon,strut=off} % DO NOT CHANGE THIS
\floatstyle{ruled}
\newfloat{listing}{tb}{lst}{}
\floatname{listing}{Listing}
\usepackage{bm}
\usepackage{array}
\usepackage{multirow}
\usepackage{makecell}
\newcommand{\tabincell}[2]{\begin{tabular}{@{}#1@{}}#2\end{tabular}}

\usepackage{hyperref}
\hypersetup{
    colorlinks=true,
    allcolors=black,
    urlcolor=blue,
}

\title{Autoregressive Sequence Modeling for 3D Medical Image Representation}
\author{
    Siwen Wang\equalcontrib \textsuperscript{\rm 3}, Churan Wang\equalcontrib \textsuperscript{\rm 2}, Fei Gao\textsuperscript{\rm 2}, Lixian Su\textsuperscript{\rm 3}, Fandong Zhang\textsuperscript{\rm 3},\\ Yizhou Wang\textsuperscript{\rm 2 \rm 4}, Yizhou Yu\thanks{Corresponding author.}\textsuperscript{\rm 1}
}
\affiliations{
    \textsuperscript{\rm 1}School of Computing and Data Science, The University of Hong Kong\\
    \textsuperscript{\rm 2}Center on Frontiers of Computing Studies, School of Computer Science,\\ Nat’l Eng. Research Center of Visual Technology, Peking University\\
    \textsuperscript{\rm 3}Deepwise AI Lab\\
    \textsuperscript{\rm 4}State Key Lab of General Artificial Intelligence, Inst. for Artificial Intelligence, Peking University\\
    wangsiwendut@gmail.com, churanwang@pku.edu.cn, yizhouy@acm.org
}

\usepackage{bibentry}
\begin{document}

\maketitle

\begin{abstract}
Three-dimensional (3D) medical images, such as Computed Tomography (CT) and Magnetic Resonance Imaging (MRI), are essential for clinical applications. However, the need for diverse and comprehensive representations is particularly pronounced when considering the variability across different organs, diagnostic tasks, and imaging modalities. How to effectively interpret the intricate contextual information and extract meaningful insights from these images remains an open challenge to the community. While current self-supervised learning methods have shown potential, they often consider an image as a whole thereby overlooking the extensive, complex relationships among local regions from one or multiple images.
In this work, we introduce a pioneering method for learning 3D medical image representations through an autoregressive pre-training framework. Our approach sequences various 3D medical images based on spatial, contrast, and semantic correlations, treating them as interconnected visual tokens within a token sequence. By employing an autoregressive sequence modeling task, we predict the next visual token in the sequence, which allows our model to deeply understand and integrate the contextual information inherent in 3D medical images. Additionally, we implement a random startup strategy to avoid overestimating token relationships and to enhance the robustness of learning. The effectiveness of our approach is demonstrated by the superior performance over others on nine downstream tasks in public datasets. 
Code is publicly available at \url{https://github.com/serena9525/AR-SSL4M}.

\end{abstract}

\section{Introduction}

The realm of medical imaging has witnessed significant evolution with the advent of advanced modalities such as Computed Tomography (CT) and Magnetic Resonance Imaging (MRI)~\cite{najjar2023redefining}. These three-dimensional (3D) medical images serve as a cornerstone for clinical diagnosis and treatment planning, providing physicians with a detailed glimpse into the inner workings of the human body~\cite{panayides2020ai}. Despite their widespread impact, the complexity and richness of 3D medical images present unique challenges in interpretation and analysis. The heterogeneity across different organs, the variability in diagnostic tasks, and the diversity of imaging modalities further complicate the comprehensive representation of these images~\cite{zhou2023nnformer}.

In the data-driven paradigm of deep learning, leveraging large-scale well-annotated data can lead to effective representation learning~\cite{ye2023uniseg,tian2024fairseg}. However, for medical image analysis, obtaining labeled data is particularly challenging, due to the intrusive nature of certain imaging modalities and the laborious process of annotation~\cite{tajbakhsh2020embracing, jin2023label}. Therefore, we consider learning a general and effective representation from large-scale unlabeled data first. This allows our models to adapt to various downstream tasks with only a small amount of labeled data, significantly reducing the dependency on large-scale annotated data. 

Moreover, recent advances in self-supervised learning (SSL) have shown promising results in visual representation learning for certain tasks~\cite{oquab2023dinov2}. These methods capitalize on the idea of reconstructing masked input data or contrastive learning to learn robust feature representations without the need for explicit labels. However, 3D medical images offer depth and volume, and are often sparse, posing new challenges when applying these methods to 3D medical image representations. 
As most existing self-supervised methods consider each image as a whole~\cite{zhou2023unified,el2024scalable,gao2024cross}, often overlooking inner and inter-correlations of 3D medical images, e.g., complex relationships among patches, modalities, and semantics. 
This limitation highlights the need for a more holistic and interconnected representation learning framework.

In this work, we address this challenge by introducing a novel self-supervised method to learn generalizable 3D medical image representations. We design a set of rules to transform diverse 3D medical images into coherent patch sequences. Every patch sequence is composed of several patches cropped from one or multiple original images according to spatial, semantic, or contrast correlation within 3D medical data. Every patch in a patch sequence is further divided into several visual tokens. All tokens from the patches in a patch sequence are concatenated to form a longer token sequence. We further introduce an autoregressive sequence modeling task to guide the network in learning the spatial, contrast, and semantic correlations among tokens. Thereby our model fosters a deep understanding and integration of the contextual information encapsulated within 3D medical images. To avoid overestimating the relationships among tokens and for better downstream adaptation, we further employ a random startup strategy. This extends the concept of prefix causal attention from individual 2D natural images~\cite{el2024scalable} to 3D medical data. This strategy randomizes the starting point of a token sequence, preventing the model from relying on consistent sequence lengths and encouraging more robust learning of the intrinsic correlations within 3D medical data. 

To summarize, our contributions are mainly three-fold:

1) We introduce a unified perspective by serializing diverse 3D medical images into token sequences. We propose an autoregressive sequence modeling task that guides the network to effectively learn generalizable 3D medical representations in a self-supervised manner.

2) We also extend prefix causal attention with random startup from 2D images to 3D medical representation to avoid overestimation of correlations among tokens and for better downstream adaptation.

3) We evaluate our method through nine downstream tasks in public datasets, such as segmentation of organs and tumours in CT or MRI and classification of COVID-19 and lung nodules. Our method achieves around 2.1\% performance improvements on segmentation and 4\%-6\% performance improvements on classification, highlighting the potential of our method to advance the field of 3D medical image analysis.

\section{Related Work}

\textbf{3D Medical Images Analysis.} Compared to 2D images, 3D medical images possess a significantly higher spatial complexity, hence conventional 2D methods are inadequate for learning 3D representation. Numerous studies have conducted meaningful explorations into 3D medical image analysis, primarily divided into methods based on 2D and 3D models~\cite{ni2019elastic,yang2021reinventing}. The advantage of 2D models lies in their ability to leverage vast amounts of natural images to obtain powerful pre-trained models. They can take 3D image planes as input channels, blending multiplanar data into 2D models~\cite{moeskops2016deep,prasoon2013deep}, or use three adjacent slices as channels~\cite{ding2017accurate,yu2018recurrent}. However, these methods are incapable of learning the context of the 3D space, which is a critical aspect for accurate representation and analysis in 3D medical images.

In contrast, 3D models are capable of learning more complex spatial features~\cite{cciccek20163d,roth2018application}. However, it is often challenging to obtain 3D models pre-trained on large-scale data. Our proposed approach obtains powerful pre-trained 3D models by capturing inner-correlations and inter-correlations of 3D medical images.

\noindent\textbf{Self-Supervised Learning.} Self-supervised learning (SSL) has emerged as a promising approach in recent years to harness unlabeled data. Initially, SSL has achieved remarkable results in the domain of natural images. The current methods are divided into two categories approximately: contrastive-based and reconstruction-based~\cite{chen2020simple, he2022masked}. The core idea of contrastive learning is to minimize the feature distance between different views of the same image while maximizing the distance between different images, thereby forcing the model to learn discriminative features for instances~\cite{chen2020simple,chen2021empirical,caron2021emerging}. Contrastive methods primarily focus on global representations and lack local focus, which leads to suboptimal performance in dense prediction tasks~\cite{zhou2021preservational, zhang2022leverage}. Reconstruction-based methods primarily focus on masked image modeling, where a significant portion of the image content is masked and then reconstructed~\cite{he2022masked, xie2022simmim}. However, they only explore representation in 2D individual nature images without considering the unique correlation within 3D medical images. 

Due to the scarcity of annotated data, SSL has also garnered significant attention in the field of medical image analysis. Firstly, the researchers have designed many proxy tasks that focus on the characteristics of medical images, particularly for 3D images~\cite{zhou2021models,tang2022self}. Additionally, based on the contrastive approach, researchers have proposed improvement strategies, as well as methods that integrate multiple proxy tasks~\cite{zhou2021preservational,zhou2023unified}. 
Although many designs have been proposed specifically for 3D medical images, current methods are often focused on individual images. There is a relative scarcity of approaches that utilize cross-sequence characteristics in 3D medical images for self-supervised representation learning. 
Taking the unique sequential correlation inherent in 3D medical images into full consideration, we propose to transform 3D images into various types of sequences. We design the autoregressive sequence modeling approach to learn inner and inter-correlation within 3D medical images.

\section{Methodology}

\begin{figure*}[t]
\begin{center}
    \includegraphics[height=0.67\linewidth]{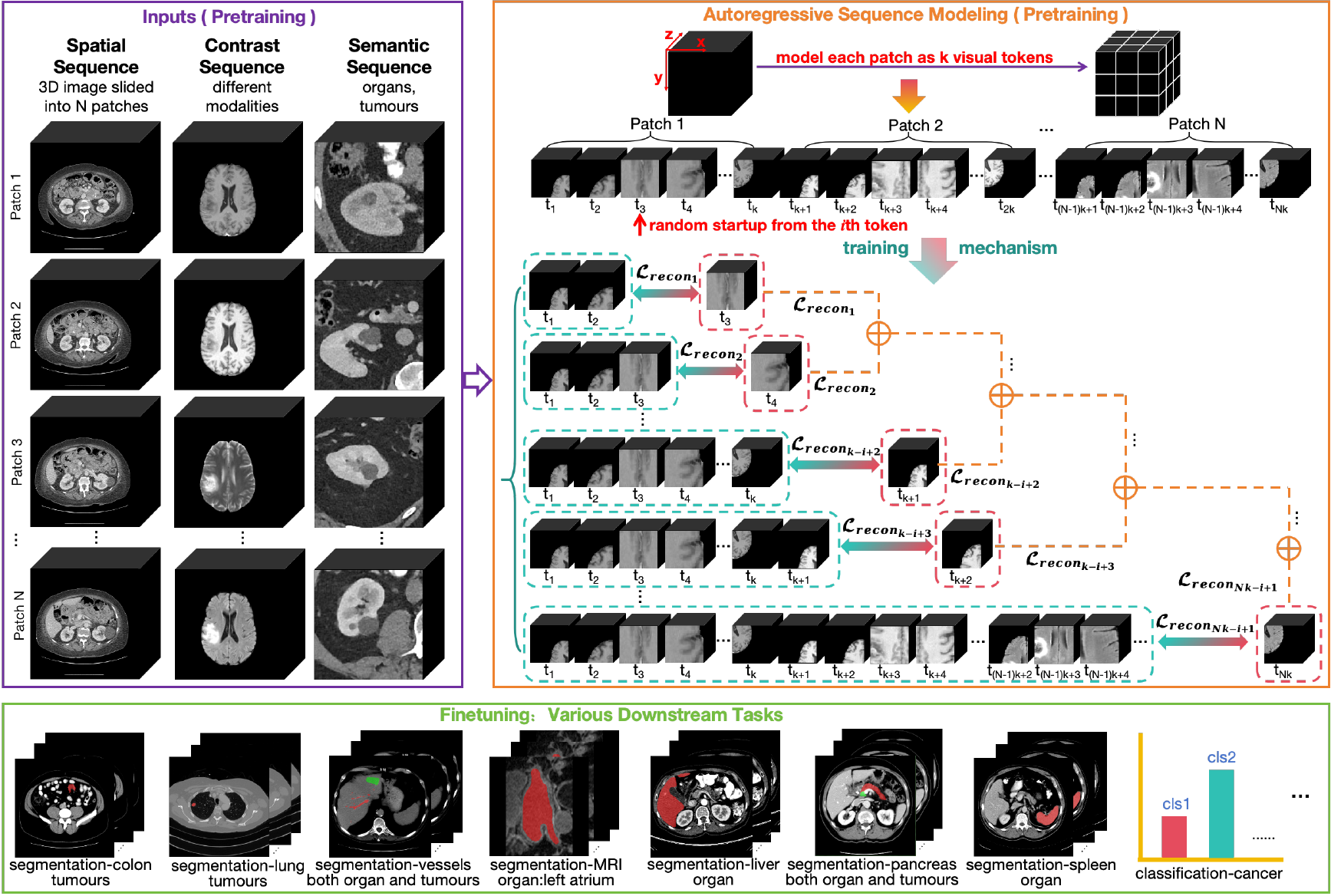}
\end{center}
    \centering
    %\vspace{-0.2cm}
    \centering\caption{Overview of our Autoregressive Sequence Modeling approach for 3D Medical Images. The left purple box shows the transformation of one or more 3D medical images into a patch sequence with N patches, highlighting spatial, contrast, and semantic relationships within 3D data. In the orange box on the right, patches within the sequence are divided into visual tokens, which are then concatenated to form an ordered token sequence. 
    During pre-training, the start of the token sequence $t_i$ is selected randomly to enhance learning robustness.
    At the bottom of the orange box, the schematic diagrams of the training mechanism demonstrate how our method leverages autoregressive modeling to predict subsequent tokens and integrate contextual information. The green box shows our method can be generalized to various downstream tasks in the fine-tuning stage.
}
\label{fig:overview}
\vspace{-0.2cm}
\end{figure*}

Our goal is to learn generalizable 3D medical image representations that can be applied across various downstream clinical tasks. As shown in Figure~\ref{fig:overview}, we start by transforming diverse 3D medical images into a set of patch sequences, capturing the inherent spatial, contrast, and semantic correlations during pre-training. We then describe our training mechanism, which involves tokenizing these sequences and employing autoregressive sequence modeling to deeply integrate contextual information. After pre-training, we introduce how the learned representations are adapted to specific clinical tasks through a fine-tuning process, showcasing the versatility and applicability of our method.

\subsection{Transform the 3D Inputs into Patch Sequences}

Our approach leverages the transformation of 3D medical images into patch sequences. This strategy provides a solution to the challenge of capturing the rich interdependencies and contextual information inherent in volumetric data. By strategically transforming these images into different types of patch sequences, our model can effectively learn from the complex patterns and relationships within 3D medical data. To achieve this, we use diverse sources of medical image data, categorizing them into three distinct types of patch sequences as below:

\noindent\textbf{Spatial Sequences.} Capitalizing on the inherent spatial relationships encoded in 3D medical images, we employ a sliding window technique (\textit{e.g.}, stride of 8). This allows creating a set of sequences that implicitly encode spatial contexts by extracting overlapping patches from 3D medical images.

\noindent\textbf{Contrast Sequences.} To leverage multi-modal images with varying contrasts (e.g. different image modalities within an MRI scan), we construct sequences that utilize these differences. By aligning and concatenating patches from the same anatomical location across different image modalities, we create sequences that capture the unique information imparted by each contrast.

\noindent\textbf{Semantic Sequences.} Given the sparsity and specialized characteristics of medical images, we use categories from the public DeepLesion~\cite{yan2018deeplesion} dataset to form semantic sequences. By grouping patches of similar semantic content, such as those from the same lesion or organ category, we enable our model to develop a robust understanding of semantic features.

\subsection{Learning by Autoregressive Sequence Modeling}
The core idea of our methodology lies in autoregressive sequence modeling, which leverages the sequential structure of our preprocessed 3D medical inputs. This section details how we use this model to learn rich, contextual representations of 3D medical data.

\noindent\textbf{Token Sequences.} As mentioned above, we first transform 3D images into patch sequences. Suppose each patch sequence has $N$ patches. The next step involves tokenizing each patch into $k$ visual tokens. All $N\times k$ tokens from a patch sequence are concatenated into a token sequence $S$ following the order mentioned above as $\mbox{T}=\{\bm{t}_{1}, ..., \bm{t}_{Nk}\}$. 

\noindent\textbf{Autoregressive Prediction.} With token sequences, we employ an autoregressive sequence model to predict the probability distribution of each token in the sequence conditioned on its predecessors. The sequence is decomposed into a product of conditional probabilities for each token, 

\begin{equation}
{P(t)}= \prod_{m=1}^{Nk} P(t_{m} \, | \, t_{<m}),
\label{eq1}
\end{equation}

where the $P(t_m|t_{<m})$ represents the probability of the $m$th token given all previous tokens in the sequence.

\noindent\textbf{Training Mechanism.} For training our model, we use a normalized pixel-level regression loss, inspired by He \textit{et al.}~\cite{he2022masked}. The loss function is designed to minimize squared distances between predicted tokens and their ground-truth values, encouraging the model to accurately predict the next token in a sequence. To align the autoregressive nature of our pre-training with bidirectional self-attention required for downstream tasks and avoid overestimating correlations among tokens, we implement a prefix self-attention mechanism. This incorporation of \textbf{a random starting token} $t_i$
ensures that all tokens preceding it are processed with bidirectional attention, capturing the comprehensive context available up to that point. This allows us to extend attention techniques used for 2D natural images~\cite{el2024scalable} to the more complex domain of 3D medical image analysis. The tokens that follow $t_i$ in the sequence are then subject to autoregressive attention, where each token's prediction is conditioned on the preceding tokens. Importantly, only the tokens subsequent to $t_i$ are included in the autoregressive prediction loss calculation, aligning with the formula:

\begin{equation}
    \label{eq2}
    \mathcal{L}_{recon} = \min \limits\frac{1}{Nk-i+1}\sum_{m=i}^{Nk} \Vert P(t_{m}) - t_{m} \Vert_2 .
\end{equation}

By training our model in this manner, our method can not only learn the intricate patterns within 3D medical images but also adapt flexibly to the diverse needs of various downstream clinical applications. Thereby the generalizability and effectiveness of our approach can be enhanced.

\subsection{Fine-tuning on Downstream Tasks}

After the pre-training stage, our model, enriched with comprehensive representations of 3D medical images, is adept at adapting to a variety of downstream tasks through a fine-tuning process. During fine-tuning, the input can be standard 3D medical images. Using the dataset for a given task, we initialize the model with pre-trained weights. We then optimize the parameters by minimizing the task-specific loss, such as Dice similarity for segmentation tasks or cross-entropy for classification. This fine-tuning approach enables rapid adaptation and enhanced performance on clinical tasks, as the model leverages the robust understanding developed during pre-training rather than starting from scratch. This strategy underscores the versatility and effectiveness of our method in advancing the analysis of 3D medical images for clinical applications.

\section{Experiments}

In this section, we first introduce the datasets used for pre-training and downstream tasks. Next, we detail the specific implementation of training and evaluation. Then, we list the methods compared in our study. Finally, we present a comprehensive set of experimental results, including a comparative analysis with existing state-of-the-art (SOTA) methods across multiple tasks, ablation studies highlighting the key components of our approach, and visualization results.

\subsection{Implementation Details} 
\subsubsection{Pre-training Datasets.} 
The pre-training datasets are divided into several sources: individual images for spatial sequences, multimodal images for contrast sequences, and images belonging to the same semantic category for semantic sequences. For individual images, we collect 23,287 3D CT and MRI volumes from 12 public medical image datasets (RibFrac~\cite{jin2020deep}, TCIA Covid19~\cite{an2020tcia}, AMOS22~\cite{ji2022amos}, ISLES2022~\cite{hernandez2022isles}, AbdomenCT-1K~\cite{ma2021abdomenct}, Totalsegmentator~\cite{wasserthal2023totalsegmentator}, Verse 2020~\cite{sekuboyina2021verse}, RSNA-2022-CSFD~\cite{flanders2022rsna}, RSNA-2020-PED~\cite{colak2021rsna}, STOIC~\cite{revel2021study}, FLARE22~\cite{ma2023unleashing}, and FLARE23~\cite{ma2024automatic}). For multimodal images, we collect 2,995 multimodal MRI scans from BraTS 23~\cite{labella2023asnr}, which is a series of challenges on brain MRI image analysis. Each scan of this dataset includes four MRI modalities (T1w, T1ce, T2w, and Flair).
Images belonging to the same semantic category are obtained from the DeepLesion dataset~\cite{yan2018deeplesion}, which contains 10,594 CT scans of 4,427 patients.

\renewcommand{\arraystretch}{1.2}

\begin{table*}[t!]
% \caption{Segmentation results on Task03 Liver, Task06 Lung, Task07 Pancreas, Task08 Hepatic Vessel, Task09 Spleen, and Task10 Colon on MSD dataset~\cite{antonelli2022medical} and LA dataset~\cite{xiong2021global}.}
\vspace{-0.2cm}
% \label{table1}
%\large
\begin{center}
\resizebox{\textwidth}{!}{
\begin{tabular}{|c|cccccc|c|c|}
\hline

\multirow{3}{*}{\textbf{Methodology}}& \multicolumn{7}{c|}{\textbf{CT: MSD dataset}} & \textbf{MRI: LA dataset} \\\cline{2-9}
& \textbf{Task03} & \textbf{Task06} & \textbf{Task07} & \textbf{Task08} & \textbf{Task09} & \textbf{Task10} & \textbf{Avg} & \textbf{Dice} \\ 
& \textbf{Liver} & \textbf{Lung} & \textbf{Pancreas} & \textbf{Hepatic Vessel} & \textbf{Spleen} & \textbf{Colon} & \textbf{Dice} & \textbf{Score} \\\hline

\textit{(Train From Scratch)}  &  &  &  &  &  & & &\\
UNETR~\cite{hatamizadeh2022unetr} & 0.9285 & 0.4758 & 0.5384 & 0.5665 & 0.9372 & 0.2446 & 0.6152 & 0.8656\\
3D UNet~\cite{cciccek20163d} & 0.9376 & 0.5222 & 0.5547 & 0.5770 & 0.9375 & 0.4057 & 0.6558 & 0.8755\\
\hline
\textit{(with General SSL)}  &  &  &  &  &  & & &\\
SimCLR~\cite{chen2020simple} & 0.9271 & 0.5631 & 0.5466 & 0.5519 & 0.9472 & 0.3330 & 0.6448 & 0.8988\\
MoCov3~\cite{chen2021empirical}  & 0.9298 & 0.5730 & 0.5563 & 0.5480	& 0.9465 & 0.4200 & 0.6623 & 0.9009\\
DINO~\cite{caron2021emerging} & 0.9392 & 0.5381 & 0.5478 & 0.5772 & 0.9470 & 0.4016 & 0.6585 & 0.9029\\
\hline
\textit{(with Medical SSL)} &  &  &  &  &  &  & &\\
PCRLv2~\cite{zhou2023unified} & 0.9451 & 0.6138 & 0.5894 & 0.5887 & 0.9417 & 0.4423 & 0.6868 & 0.9053\\
MAE3D~\cite{chen2023masked} & 0.9435 & 0.6277 & 0.5728 & 0.5878 & 0.9431 & 0.4522 & 0.6879 & 0.9054\\
MedCoSS~\cite{ye2024continual} & 0.9401 & 0.6292 & 0.5685 & 0.5890 & 0.9475 & 0.4381 & 0.6854 & 0.9062\\
\textbf{Ours} & \textbf{0.9593} & \textbf{0.6529} & \textbf{0.5910} & \textbf{0.6014} & \textbf{0.9585} & \textbf{0.4896} & \textbf{0.7088} & \textbf{0.9157}\\

\hline
\end{tabular}}
\end{center}
\vspace{-0.2cm}
\caption{Segmentation results on Task03 Liver, Task06 Lung, Task07 Pancreas, Task08 Hepatic Vessel, Task09 Spleen, and Task10 Colon on MSD dataset~\cite{antonelli2022medical} and LA dataset~\cite{xiong2021global}.}
\vspace{-0.2cm}
\label{table1}
\end{table*}

\begin{table}[h]
% \caption{Classification results of COVID-19 diagnosis on RICORD~\cite{tsai2021rsna} and of lung nodule malignancy diagnosis on LIDC-IDRI~\cite{armato2011lung}.}
% \label{table3}
\small
% \begin{tabular}{|p{3.98cm}<{\centering}|p{0.62cm}<{\centering} p{0.68cm}<{\centering}|p{0.62cm}<{\centering} p{0.68cm}<{\centering}|}
\begin{tabular}{|p{3.69cm}<{\centering}|p{0.52cm}<{\centering} p{0.68cm}<{\centering}|p{0.52cm}<{\centering} p{0.68cm}<{\centering}|}
\hline
\multirow{2}{*}{\textbf{Methodology}} & \multicolumn{2}{c|}{\textbf{COVID-19}} & \multicolumn{2}{c|}{\textbf{Lung Nodule}} \\\cline{2-5}
 & \textbf{ACC}  & \textbf{AUC}  & \textbf{ACC} & \textbf{AUC}\\
\hline
\textit{(Train From Scratch)} &  &  &  & \\
ResNet~\cite{he2016deep} & 0.7500 & 0.8133& 0.8440 & 0.8630 \\
% ViT-B~\cite{dosovitskiy2020image} & 0.7381 & 0.7940 & 0.8290 & 0.8587 \\
% \hline
ViT~\cite{dosovitskiy2020image} & 0.7381 & 0.7940 & 0.8290 & 0.8587 \\
\hline
\textit{(with General SSL)} & & & & \\
SimCLR~\cite{chen2020simple} & 0.7976 & 0.7904 & 0.8484 & 0.8605 \\
MoCov3~\cite{chen2021empirical} & 0.7738 & 0.8446 & 0.8452 & 0.8683 \\
DINO~\cite{caron2021emerging} & 0.7857 & 0.8297 & 0.8323 & 0.8755 \\
\hline
\textit{(with Medical SSL)} & & & &\\
PCRLv2~\cite{zhou2023unified} & 0.8095 & 0.8632 & 0.8516 & 0.8981 \\
MAE3D~\cite{chen2023masked} & 0.8095 & 0.8703 & 0.8419 & 0.8906 \\
MedCoSS~\cite{ye2024continual} & 0.8333 & 0.8803 & 0.8323 & 0.8983 \\
\textbf{Ours} & \textbf{0.8929} & \textbf{0.9259} & \textbf{0.8871} & \textbf{0.9361}\\
\hline
\end{tabular}
% \vspace{-0.5cm}
\caption{Classification results of COVID-19 diagnosis on RICORD~\cite{tsai2021rsna} and of lung nodule malignancy diagnosis on LIDC-IDRI~\cite{armato2011lung}.}
\vspace{-0.4cm}
\label{table3}
\end{table}

\begin{table}[h]
% \caption{Results on segmentation (LA dataset~\cite{xiong2021global}) and classification (RICORD~\cite{tsai2021rsna}) tasks under our proposed method by training with different amounts of annotated data in fine-tuning stage.}
\centering
\small
% \resizebox{\textwidth}{0.5}{
\begin{tabular}{|c|c|c|}
\hline
\textbf{Labeling Ratio} & \textbf{Dice (LA dataset)} & \textbf{AUC (RICORD)} \\\hline
100\% & 0.9157 & 0.9259 \\
50\% & 0.9039 & 0.8696\\
25\% & 0.8741 & 0.8382\\
10\% & 0.8365 & 0.8054\\
\hline
\end{tabular}

% \label{tab-labelrate}
% \vspace{-0.4cm}
\caption{Results on segmentation (LA dataset~\cite{xiong2021global}) and classification (RICORD~\cite{tsai2021rsna}) tasks under our proposed method by training with different amounts of annotated data in fine-tuning stage.}
\vspace{-0.4cm}
\label{tab-labelrate}
\end{table}

\subsubsection{Downstream Datasets.} We conducted downstream experiments in nine clinical tasks on public medical image datasets to evaluate the effectiveness of our method. These datasets cover a variety of organs, lesions, and modalities, including Task03 Liver (131 cases), Task06 Lung (64 cases), Task07 Pancreas (282 cases), Task08 Hepatic Vessel (303 cases), Task09 Spleen (41 cases), and Task10 Colon (126 cases) from Medical Segmentation Decathlon (MSD)~\cite{antonelli2022medical}, Left Atrium (LA)~\cite{xiong2021global} (100 cases), RICORD~\cite{tsai2021rsna} (330 cases) and LIDC-IDRI~\cite{armato2011lung} (1633 cases). These datasets can be categorized into 3D segmentation and 3D classification tasks. Specifically, Task03, Task09, and LA are used for organ segmentation, while Task06 and Task10 focus on tumour segmentation. Task07 and Task08 are designed for segmenting both organs and tumours. RICORD is used for COVID-19 binary classification (being COVID-19 or not). {LIDC-IDRI is used for lung nodule binary classification (level 1/2 into negative class and 4/5 into positive class, ignoring the cases with malignancy level 3) similar to other researches that have used this dataset~\cite{wu2018joint}. We randomly split the whole set into training, validation, and test at a ratio of 7:1:2 for the tasks on the MSD dataset. For the LA, RICORD, and LIDC-IDRI datasets, we follow the data split in ~\cite{yu2019uncertainty}, ~\cite{ye2024continual}, and ~\cite{yang2023medmnist}, respectively.

\subsubsection{Training and Evaluation Details} 
We use the AdamW optimizer and cosine learning rate decay scheduler for both pre-training and downstream tasks. In the pre-training stage, the initial learning rate is 1e-4, and we set 100K training steps with a batch size of 288. During the fine-tuning stage, the layer-wise learning rate decay strategy with the ratio of 0.75 is adopted for stabilizing the ViT training. 
The evaluation metric in classification tasks is the area under the receiver operator curve (AUC), and accuracy (ACC). For segmentation tasks, we use Dice similarity as the evaluation metric. To make a fair comparison, ViT-B~\cite{dosovitskiy2020image} is adopted as the backbone network, and UNETR~\cite{hatamizadeh2022unetr} is employed for segmentation tasks. More details of our implementation are provided in the supplemental material. 

\subsubsection{Compared Baselines} 
Our baselines for segmentation include 1)UNETR~\cite{hatamizadeh2022unetr} and 2) 3D UNet~\cite{cciccek20163d}, for classification include 1) ResNet~\cite{he2016deep} and 2) ViT-B~\cite{dosovitskiy2020image}, which are trained from scratch. We further compare with three general SSL methods: 1) SimCLR~\cite{chen2020simple}, 2) MoCov3~\cite{chen2021empirical}, 3) DINO~\cite{caron2021emerging}. The first two are based on contrastive learning. And DINO~\cite{caron2021emerging} introduces a self-distillation framework where a student model learns to predict the output of a teacher model. Then, we compare with three powerful methods specifically tailored for medical images: 1) PCRLv2~\cite{zhou2023unified}, 2) MAE3D~\cite{chen2023masked}, 3) MedCoSS~\cite{ye2024continual}. Specifically, MAE3D~\cite{chen2023masked} is a generative method, which leverages an autoencoder and mask-based pretext task to learn visual representations from medical images. PCRLv2~\cite{zhou2023unified} integrates pixel restoration and hybrid feature contrast into a multi-task optimization problem. MedCoSS~\cite{ye2024continual} adopts a sequential pre-training paradigm using a continual learning approach.

\begin{table*}[t!]
% \caption{Ablation study of our proposed method on segmentation (MSD Task06~\cite{antonelli2022medical}) and classification (RICORD~\cite{tsai2021rsna}) tasks.}
% \label{table_ablation}
{
\centering
\large
% \resizebox{\textwidth}{0.5}{
\resizebox{\linewidth}{!}{
\begin{tabular}{|c|c|c|c|c|c|c|c|}
\hline
\multirow{2}{*}{\textbf{Transform 3D Images into Patch Sequences}} & \multirow{2}{*}{\textbf{Proposed Training Mechanism}} & \multicolumn{3}{c|}{\textbf{Source Inputs for Pre-training}} & \multirow{2}{*}{\textbf{Stride}} & \textbf{MSD Task06} & \textbf{RICORD} \\\cline{3-5}
 & & Spatial & Contrast & Semantic & & \textbf{Dice} & \textbf{AUC} \\\hline
$\times$ & $\surd$ & $\surd$ & $\surd$ & $\surd$ & - & 0.5880 & 0.8617 \\
$\surd$ & $\times$ & $\surd$ & $\surd$ & $\surd$ & 8 & 0.5852 & 0.8389 \\
$\surd$ & $\surd$ & $\times$ & $\surd$ & $\surd$ & 8 & 0.6165 & 0.8610 \\
$\surd$ & $\surd$ & $\surd$ & $\times$ & $\surd$ & 8 & 0.6076 & 0.8710 \\ 
$\surd$ & $\surd$ & $\surd$ & $\surd$ & $\times$ & 8 & 0.6137 & 0.8632 \\
$\surd$ & $\surd$ & $\surd$ & $\surd$ & $\surd$ & 4 & 0.6258 & 0.8931\\
$\surd$ & $\surd$ & $\surd$ & $\surd$ & $\surd$ & 12 & 0.6435 & 0.8988\\
$\surd$ & $\surd$ & $\surd$ & $\surd$ & $\surd$ & 8 & \textbf{0.6529} & \textbf{0.9259} \\

\hline
\end{tabular}}}
\caption{Ablation study of our proposed method on segmentation (MSD Task06~\cite{antonelli2022medical}) and classification (RICORD~\cite{tsai2021rsna}) tasks.}
\label{table_ablation}
\end{table*}

\begin{figure*}[t]
\begin{center}
    \includegraphics[height=0.75\linewidth]{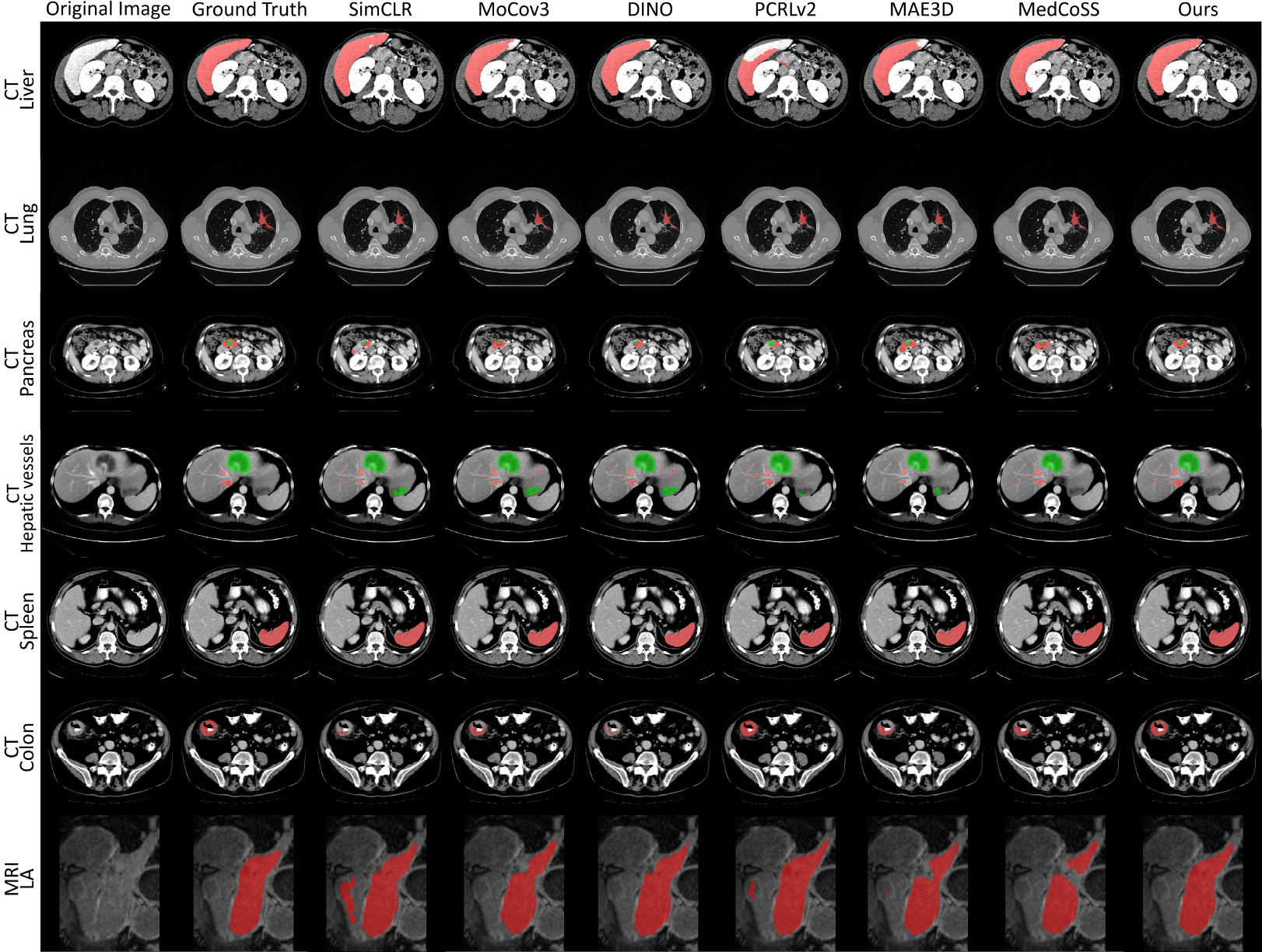}
\end{center}
    \centering
    \vspace{-0.2cm}
    \centering\caption{Visualizations of the segmentation results for various organs and pathologies from CT scans and MRI based on our proposed method and compared baselines. Each row denotes a different task. Each column denotes a different method.
}
\label{vis}
\vspace{-0.2cm}
\end{figure*}

\subsection{Results on Downstream Tasks}
We evaluate our proposed method in nine downstream tasks (organ segmentation for liver, spleen and left atrium, tumour segmentation for lung and colon, both organ and tumour segmentation for pancreas and hepatic vessel as shown in Table~\ref{table1} and COVID-19 and lung nodule malignancy binary classification as shown in Table~\ref{table3}). In addition to the different goals of the tasks themselves, there are also two different modalities in them, \textit{i.e.}, CT and MRI. 

\subsubsection{Can the learned representation be used for CT analysis?}
We first conduct comparative evaluations of the proposed approach against the SOTA methods on six CT-based segmentation tasks from the MSD challenge. As illustrated in Column2-8 of Table~\ref{table1}, our method demonstrates superior performance across all evaluated tasks on the MSD challenge, outperforming other advanced methods. The average Dice score of our method is 0.7088, which is notably higher than the scores achieved by SimCLR~\cite{chen2020simple} and MoCov3~\cite{chen2021empirical}, which are 0.6448 and 0.6623, respectively. Compared with DINO~\cite{caron2021emerging}, our method also outperforms it by a large margin. These methods are designed for general SSL, relying heavily on the negative sampling process or data augmentations, which is not well-suited for the nuanced and complex nature of medical images. We further conduct experimental comparisons with PCRLv2~\cite{zhou2023unified}, MAE3D~\cite{chen2023masked}, and MedCoSS~\cite{ye2024continual}, which employ medical SSL. However, their performance gains are still limited. Our method gains an average 2.1\% improvement in Dice score relative to the medical SSL methods. The improvements suggest that the proposed method can effectively leverage the contextual relationships inherent in 3D medical image data. 

To evaluate the performance of our method on classification tasks, we conduct experiments on the COVID-19 (RICORD) dataset and lung nodule (LIDC-IDRI) dataset. As detailed in Table~\ref{table3}, the proposed method significantly surpasses all other methods. Compared with the SOTA methods, we achieve around 4\%-6\% improvement in performance. These results suggest that our method is effective for classification tasks.

\subsubsection{Can the learned representation be used for MRI analysis?}

To validate the generalizability of our pre-trained model across different modalities, we further proceed with evaluations on the LA dataset which contains MRI data. As shown in the last column of Table~\ref{table1}, our model achieves a Dice coefficient of 0.9157, surpassing existing SOTA methods. This demonstrates the effectiveness of our approach in multimodal generalization.

\subsubsection{Can the learned representation be still useful in less annotations?} 
We take segmentation for the left atrium in LA dataset and classification for COVID-19 in RICORD dataset as examples. As shown in Table~\ref{tab-labelrate}, our model shows competitive outcomes on both tasks, despite varying amounts of annotation availability. This indicates that our model effectively harnesses extensive unlabeled datasets to develop a versatile representation, thereby reducing reliance on fully labeled datasets. Remarkably, even when utilizing as little as 10\% of the available labeled data, our model maintains satisfactory performance on both tasks, rivaling the results of compared baselines trained on complete datasets. This capability underscores the practical applicability of our model, particularly in scenarios where annotations are limited.

\subsubsection{Visualization Results.}
We also visualize the segmentation results in Figure~\ref{vis}, providing the qualitative analysis of our performance on 3D medical images. The visualizations demonstrate the effectiveness of our autoregressive sequence modeling in capturing fine details and producing high-quality segmentations.

In summary, our method outperforms other compared methods in above all tasks. These results demonstrate the effectiveness of our method in improving the segmentation and classification of various anatomical structures and pathologies. Additionally, our method shows great potential for seamless integration into diverse clinical workflows.

\subsection{Ablation Study}
We evaluate some variant models to verify the effectiveness of each component, including transforming 3D images into patch sequences, the proposed training mechanism, and the source inputs for pre-training. Take segmentation for lung tumours in MSD dataset and classification for COVID-19 in RICORD dataset as examples. Some explanations of terms in the Table~\ref{table_ablation} are lists as below:

\noindent\textbf{Transform 3D Images into Patch Sequences:} $\times$ denotes not forming patch sequences from 3D images~\cite{el2024scalable}, \textit{i.e.}, inputs of the pre-training stage are individual images. $\surd$ denotes using our proposed method, \textit{i.e.}, inputs of the pre-training stage are patch sequences, each of which is constructed from one or multiple 3D images.

\noindent\textbf{Proposed Training Mechanism:} $\times$ denotes not using our proposed training mechanism, \textit{i.e.}, capturing the correlations among tokens by using the standard autoregressive attention mechanism without random startup during pre-training.
$\surd$ denotes using our proposed method, \textit{i.e.}, capturing the correlations among tokens by using the proposed training mechanism with a random startup during pre-training.

\noindent\textbf{Source Inputs for Pre-training:} the source data used as inputs during pre-training.

\noindent\textbf{Stride:} used stride while transforming into spatial sequences. - denotes no stride operation is required if using individual images (without using our proposed transformation for patch sequences) as inputs of the pre-training stage.

In Column1, comparison between Row1 and Row8 shows a marked improvement when transforming 3D images into patch sequences is applied. This suggests that the patch sequence approach can better capture the inner and inter-correlation of the 3D medical images, e.g., the complex relationships between patches, modalities, and semantics. Row2 and Row8 in Column2 reflect the effectiveness of our proposed training mechanism with a random startup. The source of the inputs used during the pre-training phase is indicated in Column3-5 (Row3-5 and Row8). The results underscore the benefit of using a diverse set of data sources, as it enriches the model's understanding of different anatomical structures and imaging modalities. We further compare the model performance across different sliding stride settings in Column6 (Row6-8). Generally, larger sliding strides yield better performance (stride = 8 and stride = 12 outperform stride = 4). However, excessively large strides do not provide additional improvements. As shown in Table~\ref{table_ablation}, with stride = 8, the proposed method achieves the best performance on segmentation and classification tasks. The ablative results in Table~\ref{table_ablation} show that removing or changing any of the components would lead to a descent in performance.

\section{Conclusions}

In this paper, we introduce an autoregressive sequence modeling approach to 3D medical image analysis, effectively capturing the generalizable representation of 3D medical images. Our proposed model’s state-of-the-art performance across various diverse downstream tasks demonstrates its robustness and generalizability. Through a strategic fine-tuning process, our method rapidly adapts to specific clinical applications, significantly enhancing the capabilities of 3D medical image analysis and laying a solid foundation for future innovations in the field.

\section{Acknowledgements}

This work was partially supported by Hong Kong Research Grants Council under Collaborative Research Fund (Project No. HKU C7004-22G). 

\bibliography{aaai25}

\clearpage
\setcounter{page}{0}
\setcounter{table}{0}
\appendix

\twocolumn[
\begin{@twocolumnfalse}
	\section*{\centering{{\\[45pt] \LARGE Autoregressive Sequence Modeling for 3D Medical Image Representation} ~\\ 
    {\LARGE (Supplementary Materials)\\[62pt]}}}
\end{@twocolumnfalse}
]

\subsection{Implementation Details}
\subsection{Pre-training datasets}
We provide more details of our pre-training datasets as shown in Table~\ref{tables1}.

\noindent \textbf{For Spatial Sequences.}
Spatial Sequences mainly aim at exploring the spatial correlation within 3D medical images. For spatial sequences, we collect 23,287 3D CT and MRI scans from 12 public medical image datasets, as shown in table~\ref{tables1}. 
To form a spatial sequence, we first randomly crop four 32$\times$128$\times$128 patches from the CT or MRI volume by sliding along the depth dimension with a stride of 8. These patches are then concatenated into a spatial sequence of four patches. We construct 500k spatial sequences in total. 

\noindent \textbf{For Contrast Sequences.}
Contrast Sequences mainly aim at exploring correlation across multimodalities within 3D medical images. Thus, for contrast sequences, we collect 2,995 multimodal MRI scans from BraTS2023, as shown in Table~\ref{tables1}. Each scan in this dataset includes four modalities (T1w, T1ce, T2w, and Flair). Contrast sequences are constructed from the aligned multimodal images. Specifically, we randomly crop four 32$\times$128$\times$128 patches at the same anatomical location from the T1w, T1ce, T2w, and Flair modalities, respectively. Then, we concatenate these patches into a contrast sequence of four patches. In total, we construct 300k contrast sequences.

\noindent \textbf{For Semantic Sequences.} 
Semantic Sequences mainly aim at exploring correlation across medical categories and domain semantics (e.g., lesion types, organ types). Thus, each patch in a semantic sequence is constructed by key slices of the target semantics along with its context slices, which are 30mm of extra slices around the key slices. Patches are resized into 32$\times$128$\times$128, and are concatenated into a semantic sequence of four patches with the same semantic category. Selected semantic sequences contain various lesions or organs, such as bone, abdomen, mediastinum, liver, lung, kidney, soft tissue, and pelvis respectively. Samples for semantic sequences are obtained from the DeepLesion dataset, which contains 10,594 scans. We construct 160k semantic sequences in total.

\subsection{Downstream datasets}
In Table~\ref{tables2}, we demonstrate the details of the downstream tasks we use.

\noindent \textbf{MSD dataset.} 
MSD (Medical Segmentation Decathlon) ~\cite{antonelli2022medical} dataset includes six CT-based segmentation tasks from different organs, i.e., Task03 Liver, Task06 Lung, Task07 Pancreas, Task08 Hepatic Vessel, Task09 Spleen, and Task10 Colon. We follow the pre-processing steps in ~\cite{tang2022self}. The total number of samples we use is 947. 

\noindent \textbf{LA dataset.} 
LA~\cite{xiong2021global} dataset is applied to segment the left atrium (LA) from the late gadolinium-enhanced MRI imaging. We follow the pre-processing steps and dataset split in ~\cite{yu2019uncertainty}. The number of samples we use is 100. 

\noindent \textbf{RICORD dataset.} 
RICORD~\cite{tsai2021rsna} dataset is created for COVID-19 diagnosis. The dataset includes two classes: COVID-19, and normal. We follow the pre-processing steps and dataset split in \cite{ye2024continual}. The number of samples we use is 330. 

\noindent \textbf{LIDC-IDRI dataset.} 
LIDC-IDRI dataset~\cite{armato2011lung} is designed for 5-level lung nodule malignancy classification. Similar to other researches that have used this dataset~\cite{wu2018joint,hussein2017risk}, we 
use it for lung nodule binary classification (level 1/2 into negative class and 4/5 into positive class, ignoring the cases with malignancy level 3). We follow the pre-processing steps and dataset split in \cite{yang2023medmnist}. The number of samples we use is 1633.

\subsection{Training and Evaluation Details}
We conduct pre-training on 4 NVIDIA A6000 GPUs and downstream tasks on a single NVIDIA 3090 GPU. In the pre-training stage, the patch size and the sequence length are determined by balancing the richness of features in images with the computational complexity. As shown in Table~\ref{tables2}, the patch size of inputs is 96$\times$96$\times$96 for MSD dataset, 64$\times$192$\times$192 for LA dataset, 64$\times$128$\times$128 for RICORD dataset, and 128$\times$128$\times$128 for LIDC-IDRC dataset respectively. The learning rate is set to 3e-4, 3e-4, 5e-5, and 1.5e-5 for MSD, LA, RICORD, and LIDC-IDRC respectively. We use AdamW as the optimizer. The batch size is 4, 4, 8, and 16 for MSD, LA, RICORD, and LIDC-IDRC respectively. The training epochs are set to 1000 for MSD Task03, MSD Task06, MSD Task09, MSD Task10, and LA dataset, 500 for MSD Task07, and MSD Task08 dataset, 200 for RICORD dataset, and 100 for LIDC-IDRI dataset.

\renewcommand{\arraystretch}{1.2}

\begin{table*}[t]
\vspace{-200pt}
\small
\begin{center}
% \resizebox{\textwidth}{!}{
\begin{tabular}{|c|c|c|c|c|}
\hline
\multicolumn{2}{|c|}{\multirow{2}{*}{\textbf{Datasets}}} & \multicolumn{2}{c|}{\textbf{The Number of Scans}} & \textbf{The Number of} \\\cline{3-4}
 \multicolumn{2}{|c|}{} & \textbf{Each Dataset} & \textbf{Total} & \textbf{Patch Sequences}\\
\hline
\multirow{12}{*}{\tabincell{c}{Spatial \\ Sequences}} & RibFrac~\cite{jin2020deep} & 660 & \multirow{12}{*}{23,287} & \multirow{12}{*}{500k}\\
& TCIA Covid19~\cite{an2020tcia}  & 770 & &\\
& AMOS22~\cite{ji2022amos}  & 600 & &\\ 
& ISLES2022~\cite{hernandez2022isles}  & 955 & &\\
& AbdomenCT-1K~\cite{ma2021abdomenct}  & 1,112 & &\\
& Totalsegmentator~\cite{wasserthal2023totalsegmentator} & 1,206 & &\\
& Verse 2020~\cite{sekuboyina2021verse}  & 374 &  &\\
& RSNA-2022-CSFD~\cite{flanders2022rsna}  & 2,022 & &\\
& RSNA-2020-PED~\cite{colak2021rsna}  & 7,488 & &\\
& STOIC~\cite{revel2021study}   & 2,000 & &\\
& FLARE22~\cite{ma2023unleashing}  & 2,100 & &\\
& FLARE23~\cite{ma2024automatic} & 4,000 & &\\
\hline
\multirow{5}{*}{\tabincell{c}{Contrast \\Sequences}} & BraTS2023-MEN~\cite{labella2023asnr}  & 1,141 & \multirow{5}{*}{2,995} & \multirow{5}{*}{300k}\\
& BraTS2023-MET~\cite{moawad2023brain}  & 165 & &\\
& BraTS2023-PED~\cite{kazerooni2023brain} & 144 & &\\
& BraTS2023-SSA~\cite{adewole2023brain} & 75 & &\\
& BraTS2023-GLI~\cite{baid2021rsna,menze2014multimodal,bakas2017advancing} & 1,470 & &\\
\hline

\tabincell{c}{Semantic \\ Sequences} & DeepLesion~\cite{yan2018deeplesion} & 10,594 & 10,594 & 160k\\

\hline
% \end{tabular}}
\end{tabular}
\caption{The details of pre-training datasets.}
\label{tables1}
\end{center}
\end{table*}

\begin{table*}[t]
% \centering
\vspace{-410pt}
\small
\centering
\begin{tabular}{|c|c|c|c|c|}
\hline
Datasets & Tasks & Modalities & The Number of Samples & Patch Size\\\hline
MSD Task03 & Organ Segmentation (Liver) & CT & 131 & 96$\times$96$\times$96\\
MSD Task06 & Tumour Segmentation (Lung) & CT & 64 & 96$\times$96$\times$96\\
MSD Task07 & Organ and Tumour Segmentation (Pancreas) & CT & 282 & 96$\times$96$\times$96\\
MSD Task08 & Organ and Tumour Segmentation (Hepatic Vessel) & CT & 303 & 96$\times$96$\times$96\\
MSD Task09 & Organ Segmentation (Spleen) & CT & 41 & 96$\times$96$\times$96\\
MSD Task10 & Tumour Segmentation (Colon) & CT & 126 & 96$\times$96$\times$96\\
LA & Organ Segmentation (Left Atrium) & MRI & 100 & 64$\times$192$\times$192\\
RICORD & COVID-19 Classification & CT & 330 & 64$\times$128$\times$128\\
LIDC-IDRC & Lung Nodule Classification & CT & 1633 & 128$\times$128$\times$128\\

\hline
\end{tabular}
\caption{\centering{The details of downstream tasks.}}
\label{tables2}
\end{table*}

\end{document}